\documentclass[letterpaper, 10 pt, conference]{ieeeconf}  

\IEEEoverridecommandlockouts                              
\let\labelindent\relax
\let\labelindent\relax
\usepackage{graphicx} \graphicspath{ {figures/} }
\usepackage{amsmath,amssymb,mathabx,amsbsy,mathtools,etoolbox}
\usepackage[utf8]{inputenc} 
\usepackage[T1]{fontenc}    
\usepackage{url}
\usepackage{algorithmic}
\usepackage[ruled]{algorithm2e}
\usepackage{hyperref}
\hypersetup{
    colorlinks=true,
    bookmarks=true,
    bookmarksopen=false,
    bookmarksnumbered=true,
    linkcolor=magenta,
    filecolor=[rgb]{.043 0.23 .043},      
    urlcolor=magenta,
    citecolor=[rgb]{.043 0.23 .043},
}

\usepackage{acronym}
\usepackage{enumitem}
\usepackage{balance}
\usepackage{xspace}
\usepackage{setspace}
\usepackage[skip=3pt,font=small]{subcaption}
\usepackage[skip=3pt,font=small]{caption}
\usepackage[dvipsnames,svgnames,x11names]{xcolor}
\usepackage[capitalise,nameinlink]{cleveref}
\usepackage{booktabs,tabularx,colortbl,multirow,multicol,array,makecell}
\usepackage{pifont}
\usepackage{cite}
\usepackage{nicematrix}
\usepackage{dblfloatfix}

\makeatletter
\DeclareRobustCommand\onedot{\futurelet\@let@token\@onedot}
\def\@onedot{\ifx\@let@token.\else.\null\fi\xspace}

\makeatother

\makeatletter
\def\BState{\State\hskip-\ALG@thistlm}
\makeatother

\makeatletter
\renewcommand{\paragraph}{%
  \@startsection{paragraph}{4}%
  {\z@}{0ex \@plus 0ex \@minus 0ex}{-1em}%
  {\hskip\parindent\normalfont\normalsize\bfseries}%
}
\makeatother

\crefname{algorithm}{Alg.}{Algs.}
\Crefname{algocf}{Algorithm}{Algorithms}
\crefname{section}{Sec.}{Secs.}
\Crefname{section}{Section}{Sections}
\crefname{table}{Tab.}{Tabs.}
\Crefname{table}{Table}{Tables}
\crefname{figure}{Fig.}{Fig.}
\Crefname{figure}{Figure}{Figure}

\definecolor{gblue}{HTML}{4285F4}
\definecolor{gred}{HTML}{DB4437}
\definecolor{ggreen}{HTML}{0F9D58}

\definecolor{mygray}{gray}{.92}

\definecolor{lightgray}{gray}{0.9}

\acrodef{mata}[MATA]{multi-agent task allocation}
\acrodef{mrs}[MRS]{multi-robot system}
\acrodef{llm}[LLM]{Large language model}
\acrodef{dof}[DoF]{Degree of Freedom}

\title{\LARGE \bf Driving Animatronic Robot Facial Expression From Speech}

\author{Boren Li$^{*\dagger}$, Hang Li$^*$, Hangxin Liu$^{\dagger}$
\thanks{*Boren Li and Hang Li contributed equally to this work.}
\thanks{$\dagger$ Corresponding author.}
\thanks{The authors are with State Key Laboratory of General Artificial Intelligence, Beijing Institute for General Artificial Intelligence (BIGAI). Emails: \tt{\{liboren, lihang, liuhx\}@bigai.ai}}
}

\begin{document}

\maketitle

\begin{abstract}

Animatronic robots hold the promise of enabling natural human-robot interaction through lifelike facial expressions. However, generating realistic, speech-synchronized robot expressions poses significant challenges due to the complexities of facial biomechanics and the need for responsive motion synthesis. This paper introduces a novel, skinning-centric approach to drive animatronic robot facial expressions from speech input. At its core, the proposed approach employs linear blend skinning (LBS) as a unifying representation, guiding innovations in both embodiment design and motion synthesis. LBS informs the actuation topology, facilitates human expression retargeting, and enables efficient speech-driven facial motion generation. This approach demonstrates the capability to produce highly realistic facial expressions on an animatronic face in real-time at over 4000 fps on a single Nvidia RTX 4090, significantly advancing robots' ability to replicate nuanced human expressions for natural interaction. To foster further research and development in this field, the code has been made publicly available at: \url{https://github.com/library87/OpenRoboExp}.

\end{abstract}

\setstretch{0.973}
\section{introduction}\label{sec:intro}
Accurately replicating human facial expressions is crucial for natural human-robot interaction~\cite{fong2003survey, breazeal2016social, saunderson2019robots}. Recent studies have incorporated human motion transfer techniques for animatronic faces~\cite{ren2016automatic, hyung2019optimizing, chen2021smile, yang2022optimizing, tang2023automatic}, primarily focusing on mimicking observed human expressions through leader-follower mapping. However, to achieve genuine emotional resonance with humans, social robots require speech-synchronized, lifelike expressions~\cite{lazzeri2018influence, lomas2022resonance}. This necessitates a shift towards speech-driven approaches for generating dynamic and contextually appropriate facial expressions. Such advancements have significant implications for various applications, such as entertainment, education and healthcare, where natural and expressive human-robot interaction is paramount.

\begin{figure}[t!]
    \centering
    \includegraphics[width=\linewidth]{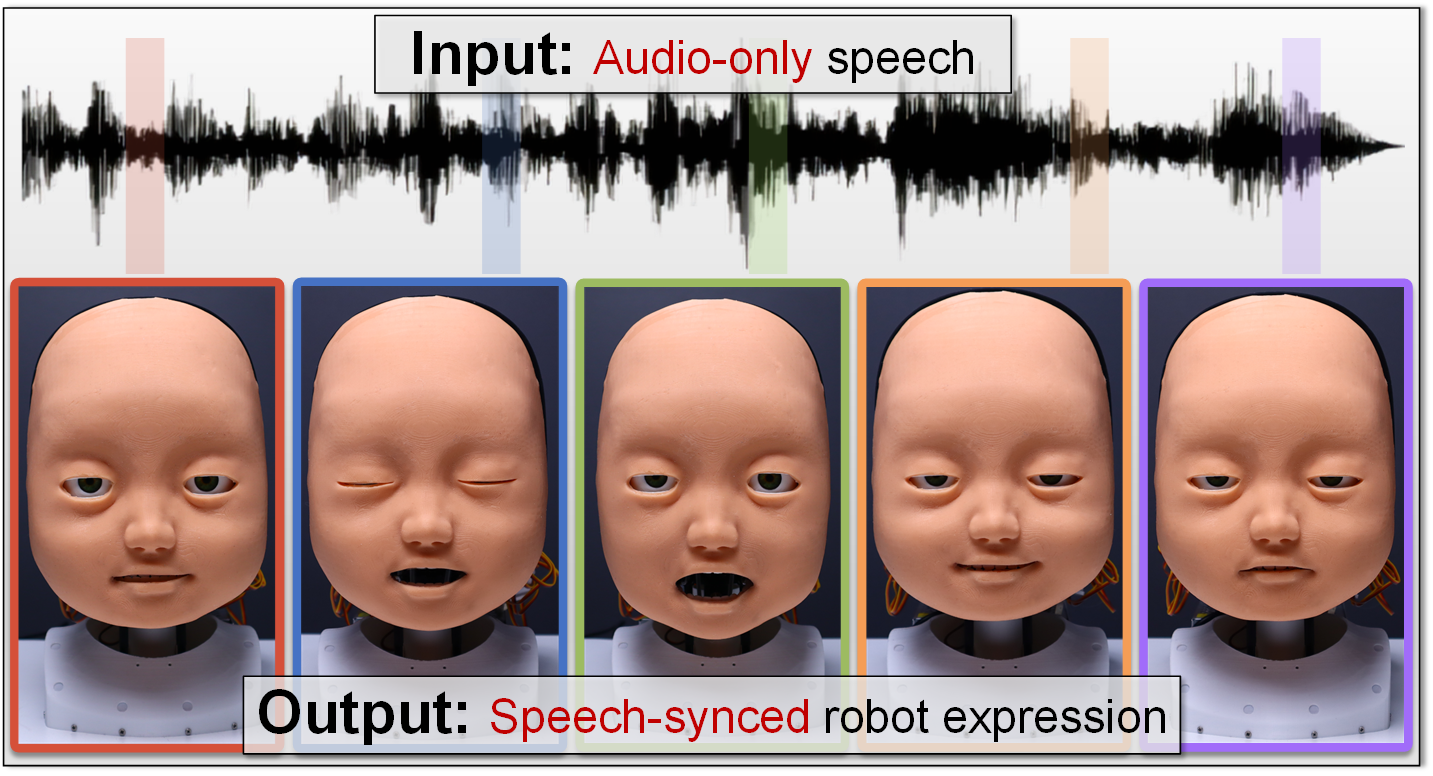}
    \caption{\textbf{Dynamic animatronic robot facial expressions generated from speech.} The figure shows the system's capability to produce diverse and lifelike facial expressions in real-time, synchronized with the corresponding audio speech input. The waveform at the top represents the audio input, while the series of images below showcase the robot's facial responses at different time points.}
    \label{fig:intro}
\end{figure}

This paper introduces the first principled approach for creating an animatronic robot face capable of generating expressions directly from speech, marking a significant advancement in creating dynamically expressive robot faces and enhancing the potential for natural human-robot interaction. Fig.~\ref{fig:intro} showcases the realism and diversity of the generated robot facial expressions, synchronized with the corresponding speech input over time.

Generating seamless, real-time animatronic facial expressions from speech presents two main challenges: (1) replicating the intricate biomechanics of human facial musculature~\cite{berns2006control, oh2006design, hashimoto2006development, mazzei2012hefes, lin2016expressional, asheber2016humanoid, faraj2021facially, yan2024facial}, and (2) generating nuanced human expressions through responsive algorithms based on advanced imitation learning~\cite{ren2016automatic, hyung2019optimizing, chen2021smile, yang2022optimizing, tang2023automatic}. Overcoming these challenges necessitates a comprehensive approach that integrates embodiment design and motion synthesis.

Conventional muscle-centric embodiment design approaches attempt to replicate human anatomy, but face significant engineering obstacles. These include the complexity of mimicking the multitude of facial muscles, their interconnections, and their subtle interactions. Additionally, the miniaturization of actuators to fit within the confined space of a robot face while maintaining the required force and precision presents considerable challenges. The key insight of this research is that achieving realistic facial skinning motions, rather than replicating internal muscle movements, is the primary objective. Consequently, this paper proposes a novel skinning-centric embodiment design approach.

Current methods animate robot facial skinning through 3D landmark alignment between humans and robots~\cite{ren2016automatic, chen2021smile}. However, this sparse landmark approach has inherent limitations: (1) insufficient capture of expression intricacies, resulting in oversimplification; (2) coupling of facial shapes and expressions, leading to inconsistencies from varying shapes; (3) topological differences hindering viable landmark motion transfer, limiting adaptability; and (4) confinement of edits to specific landmarks, restricting semantic adjustability. These limitations underscore the need for a 3D landmark-free skinning representation.

To address these challenges, this paper proposes a principled approach leveraging linear blend skinning (LBS) for both embodiment design and motion synthesis. LBS enables efficient and controllable facial expressions by linearly combining predefined shape variations (blendshapes) from a neutral shape. For embodiment, LBS guides a skinning-oriented actuation topology optimized for blend skinning objectives while referencing facial anatomy. For synthesis, LBS facilitates motion retargeting from human demonstrations into robot skinning references. A speech-driven model is further proposed that generates highly realistic, lip-synchronized LBS-based skinning motions in real-time through imitation learning. This approach significantly advances upon recent works by offering enhanced expressiveness through continuous dense deformation, improved consistency and adaptability to different facial topologies, semantic editability without need to retrain for different robot faces, and efficient real-time performance crucial for natural human-robot interaction.

In summary, this paper presents the first principled approach for creating an animatronic robot face capable of generating expressions from speech. The proposed skinning-centric approach for embodiment design and motion synthesis significantly advances the state-of-the-art in creating dynamically expressive robot faces for natural interaction.

The rest of the paper is organized as follows: Section II reviews related works in animatronic robot faces and facial expression synthesis. Section III introduces the proposed LBS-based approach. Section IV details the skinning-oriented robot development process. Section V elaborates on the skinning motion imitation learning method. Section VI presents experimental results, and Section VII concludes with a discussion of implications and future research directions.

\section{Related Works}\label{sec:related_works}
\begin{figure*}[t!]
    \centering
    \includegraphics[width=\linewidth]{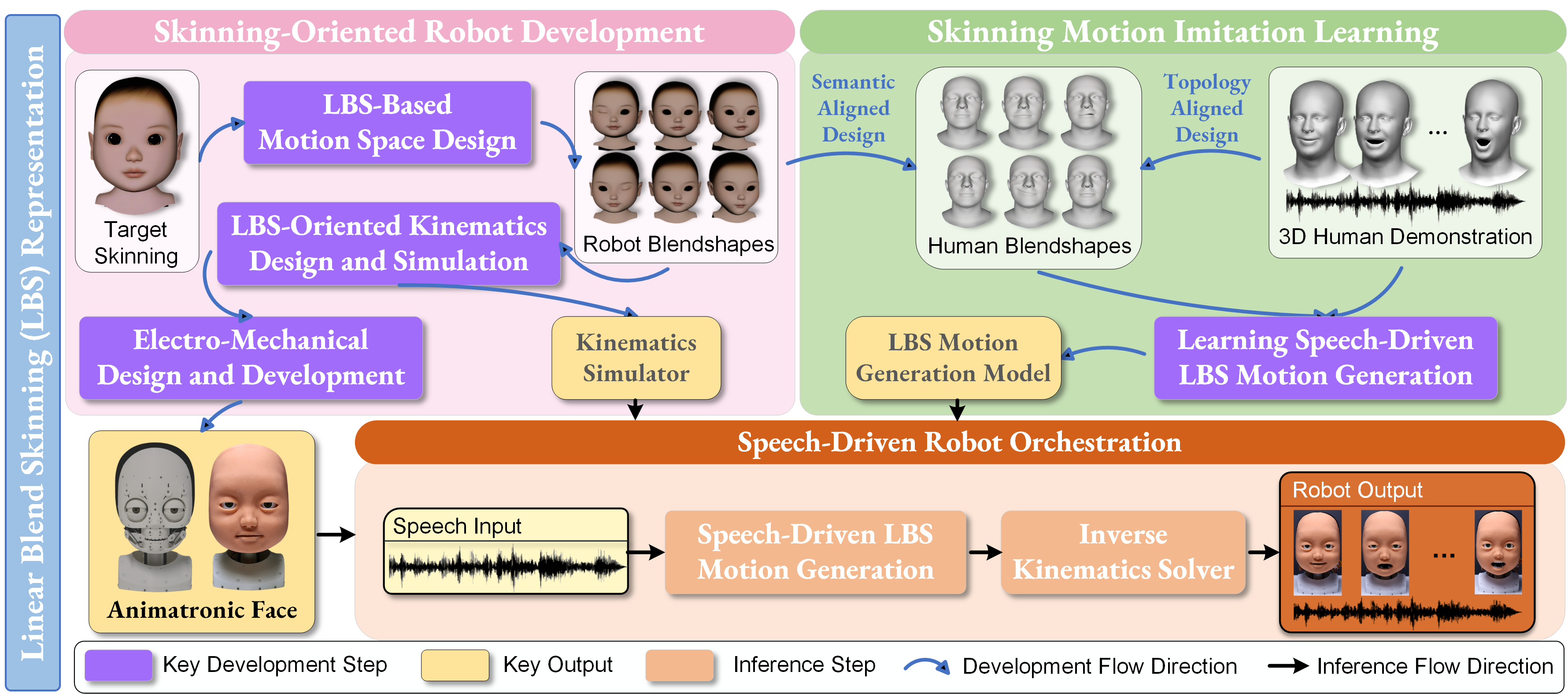}
    \caption{\textbf{The proposed approach for creating a speech-driven animatronic robot face using LBS.} The approach comprises three major components: (1) \textit{skinning-oriented robot development} designs and constructs the animatronic face paired with a kinematics simulator based on the target skinning appearance, (2) \textit{skinning motion imitation learning} involves training an LBS-based model from 3D human demonstrations to generate facial expressions from speech input, and (3) \textit{speech-driven robot orchestration} generates animatronic facial expressions during inference by utilizing the developed platform, simulator, and learned model. The diagram highlights key development steps, outputs, and inference processes, demonstrating the end-to-end workflow from concept to final animatable robot face.}
    \label{fig:approach_overview}
    \vspace{-5pt}
\end{figure*}

\textbf{Animatronic Robot Face:} The evolution of animatronic robot faces can be categorized into two main phases: (1) early approaches that prioritized hardware design with pre-programmed expressions, and (2) recent studies that incorporate human motion transfer techniques. Pioneering works~\cite{berns2006control, oh2006design, hashimoto2006development, mazzei2012hefes, lin2016expressional, asheber2016humanoid} relied heavily on pre-programmed hardware and expressions, which severely limited their ability to generalize beyond a fixed set of postures. In recent years, researchers have made significant strides by integrating human motion transfer techniques, such as active appearance models~\cite{ren2016automatic}, genetic algorithms~\cite{hyung2019optimizing}, visual mimicry learning~\cite{chen2021smile}, Bayesian optimization~\cite{yang2022optimizing}, and MAP-Elites algorithms~\cite{tang2023automatic}. While these approaches have enhanced the flexibility of animatronic robot faces, they remain fundamentally constrained by their reliance on mimicking observed human expressions through a leader-follower mapping paradigm. The creation of real-time facial expressions directly from speech on animatronic platforms remains an unexplored frontier, impeding natural human-robot interaction.

\textbf{Facial Expression Synthesis:} In parallel, speech-driven facial expression synthesis has seen substantial advancements in generating photo-realistic talking head videos~\cite{zhou2020makelttalk, liang2022expressive, zhang2023sadtalker, wang2024facecomposer}. These approaches map speech to target video domains, yielding impressive visual results. More pertinent to this work are methods for speech-driven 3D facial animation that control full vertex-level facial skinning~\cite{karras2017audio, cudeiro2019capture, richard2021meshtalk, fan2022faceformer, danvevcek2023emotional}. However, while these methods excel at creating realistic virtual renditions, they do not address the distinct challenges inherent in physical robots, such as actuation limitations, real-time control requirements, and physical constraints. Their direct applicability to animatronic platforms remains limited.

This work distinguishes itself by introducing the first principled approach for generating robot facial expressions directly from speech on real animatronic platforms. It bridges the gap between speech input and physical actuation, a critical step towards creating dynamically expressive robot faces capable of natural, speech-synchronized interactions. By addressing the unique challenges of translating speech into physical facial movements, this research opens new avenues for enhancing the expressiveness and naturalness of human-robot interaction.

\section{Proposed Approach}\label{sec:proposed_approach}

\subsection{Approach Overview}\label{subsec:approach_overview}

Fig.~\ref{fig:approach_overview} presents the proposed approach for creating a speech-driven animatronic robot face using LBS. It comprises three key components: First, skinning-oriented robot development designs and constructs the animatronic platform paired with a kinematics simulator based on the target skinning appearance. The LBS motion space is designed following blendshape design protocols~\cite{lewis2014practice} (predefined bases of facial expressions) to enable seamless motion transfer. The actuation topology and simulator are developed concurrently to match this motion space, while the robot face is constructed considering physical constraints. Second, skinning motion imitation learning involves learning an LBS-based model from 3D human demonstrations to generate robot facial motions. Human blendshapes are carefully designed to align semantically with the robot blendshapes, ensuring consistent and expressive motion transfer. The learned model takes speech as input and outputs skinning motions as reference signals. Finally, during inference, speech-driven robot orchestration generates expressions on the animatronic face using the learned model and the developed simulator. Inverse kinematics is solved online to compute actuator commands from the generated skinning motions, enabling real-time, speech-driven expressions.

\subsection{LBS Representation}\label{subsec:lbs_representation}

The proposed approach leverages LBS, a widely-used technique in computer graphics for deforming 3D meshes. LBS represents a mesh as a weighted combination of a neutral shape and a set of predefined shape variations, called blendshapes. It offers key advantages: First, it compactly encodes intricate expressions into a small set of blendshape coefficients, facilitating human-like motions with minimal data. Second, as an expression-dependent but subject-invariant representation, LBS enables consistent performance across different faces by generating the same blendshape coefficients for the same expressions. This decoupling of embodiment design from motion synthesis allows the LBS-based motion synthesis model to generalize across diverse embodiments and enables semantic editability before retargeting without retraining.

Crucially, the approach relies on semantically aligning the design of human and robot blendshapes to enable motion transfer. By carefully designing these aligned blendshapes to represent the same set of facial expressions (e.g., eye blink, mouth smile), it ensures that the same blendshape coefficients, when applied to both human and robot, produce semantically corresponding expressions, even with differing underlying facial structures.

The robot facial skinning function $T^R(\boldsymbol{\theta}):\mathbb{R}^{\boldsymbol{\theta}}\mapsto\mathbb{R}^{3U}$ represented using LBS is
\begin{equation}
    T^R(\boldsymbol{\theta}) = \mathbf{T}^R + B_E^R(\boldsymbol{\theta};\mathbf{\mathcal{E}}^R),
    \label{eq:robot_LBS}
\end{equation}
where $\mathbf{T}^R\in\mathbb{R}^{3U}$ is the neutral (zero-pose) robot face with $U$ vertices, and $B_E^R(\boldsymbol{\theta};\mathbf{\mathcal{E}}^R):\mathbb{R}^{\boldsymbol{\theta}}\mapsto\mathbb{R}^{3U}$ is the expression skinning function with $B$ robot blendshape bases $\mathbf{\mathcal{E}}^R$ and coefficients $\boldsymbol{\theta}$. The expression skinning function $B_E^R$ linearly combines the robot blendshape bases $\mathbf{\mathcal{E}}^R$ using the coefficients $\boldsymbol{\theta}$ to generate the final facial deformation, allowing efficient representation and control of diverse expressions by adjusting the blendshape coefficients.

To enable motion transfer, human blendshapes $\mathbf{\mathcal{E}}^H$ are designed to align semantically with $\mathbf{\mathcal{E}}^R$. The human mesh topology matches that of the captured 3D human demonstrations, with the mean face shape of demonstration subjects serving as the neutral human face $\mathbf{T}^H$. The human facial skinning function $T^H(\boldsymbol{\theta}):\mathbb{R}^{\boldsymbol{\theta}}\mapsto\mathbb{R}^{3V}$ is
\begin{equation}
    T^H(\boldsymbol{\theta}) = \mathbf{T}^H + B_E^H(\boldsymbol{\theta};\mathbf{\mathcal{E}}^H),
    \label{eq:human_LBS}
\end{equation}
where $V$ is the number of human face vertices. By transferring $\boldsymbol{\theta}$, human motions are effectively mapped to the robot.

For the blendshape design, the Apple ARKit standard (excluding \textit{tongue-out}) is adopted for its semantic meaningfulness and comprehensiveness, enhancing practical interoperability by leveraging existing frameworks.

\begin{figure*}[!t]
    \centering
    \includegraphics[width=\linewidth]{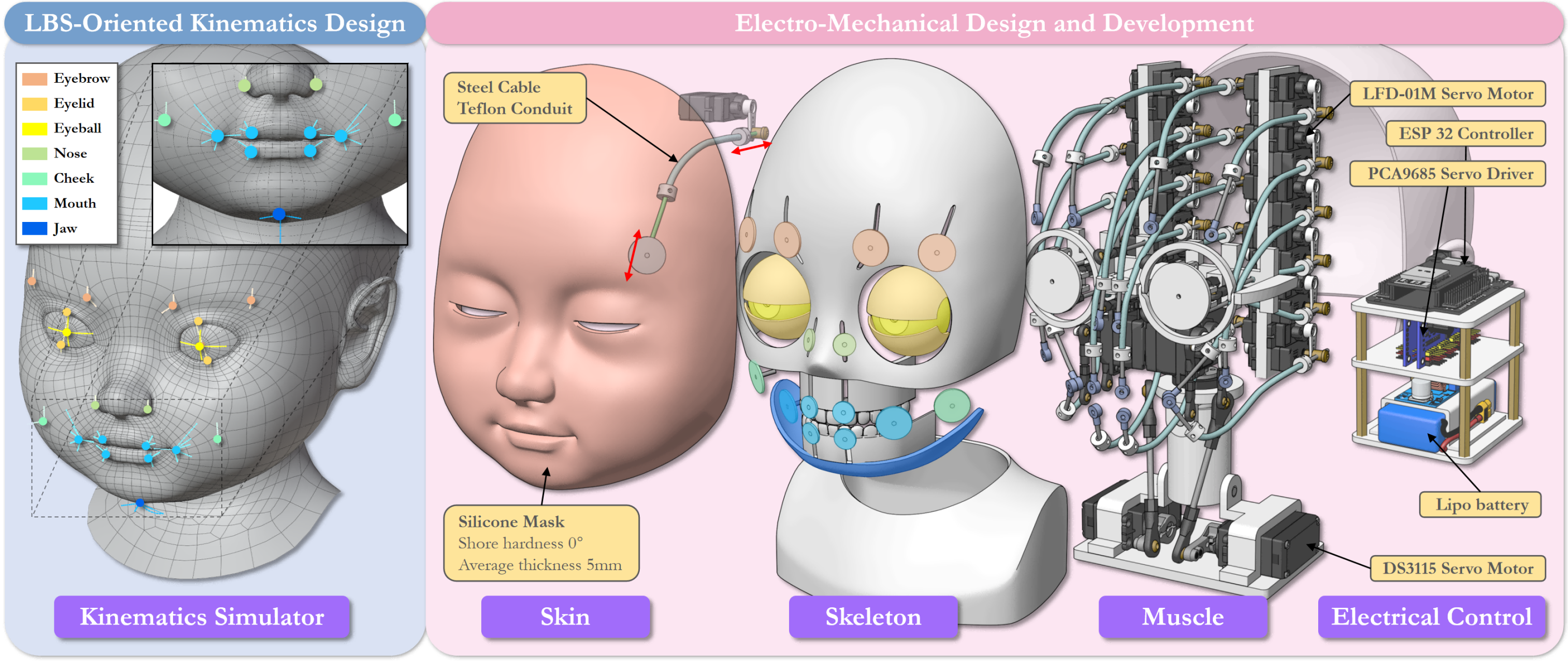}
    \caption{\textbf{The proposed skinning-oriented robot design.} The figure comprises two primary components: (1) \textit{LBS-oriented kinematics design}, which showcases the facial mesh model with strategically placed control points for various facial features to achieve actuation topology for the facial muscular system that matches the designed LBS motion space and references facial anatomy, and (2) \textit{electro-mechanical design and development} accounting for physical constraints of the embodiment, including key mechanical components of the skin, skeleton and muscular system, as well as the electrical control system. This comprehensive view demonstrates how the theoretical LBS model is translated into a functional, physically embodied animatronic face.}
    \label{fig:skinning_oriented_robot_development}
    \vspace{-10pt}
\end{figure*}

\section{Skinning-oriented Robot Development}\label{sec:skinning_oriented_robot_development}

\subsection{LBS-Oriented Kinematics Design}\label{subsec:lbs_oriented_kinematics_design}

The primary objective in designing the facial kinematics system is to reproduce the target LBS-based motion space, prioritizing functional equivalence over anatomical replication. A baby-like face serves as the target animatronic robot appearance, demonstrating the proposed skinning-centric approach's applicability to various facial morphologies.

To facilitate the design process, a kinematics simulator with adjustable topology is developed in Blender, enabling rapid computation of LBS and muscular motion spaces for iterative optimization. The design focuses on allocating skinning control points, each with bounded 6 degrees of freedom (B6DOF) for position and orientation, balancing motion flexibility and actuator complexity.

As depicted in Fig.~\ref{fig:skinning_oriented_robot_development}, the optimized kinematics design contains a total of $21$ control points: $4$ for eyebrows, $4$ for eyelids, $2$ for eyeballs, $2$ for the nose, $2$ for cheeks, $6$ for the mouth, and $1$ for the jaw. The color-coded points show the control point locations, and lines indicate the B6DOF motion bounds. This design strikes a balance between motion flexibility and system complexity.

\subsection{Electro-Mechanical Design and Development}\label{subsec:electro_mechanical_design}

\textbf{Mechanical Design:} The mechanical design aims to physically realize the facial muscular motion space defined by the control points while accounting for physical constraints. A tendon-driven actuation approach, inspired by human facial anatomy, enables the remote location of actuators, providing power under spatial limitations.

Specifically, $24$ actuators drive the eyes, eyebrows, nose, cheeks, and mouth, using tendons to actuate control points. Each actuator pulls or pushes a $1.5\mathrm{mm}$ steel cable through a $1.5\mathrm{mm}$ inner diameter Teflon conduit, enabling precise displacement control of rocker arms. Additionally, four actuators and a ball-joint linkage control the jaw module for articulation and 3D translation. The neck module, with three actuators, enables head roll, pitch, and yaw.

The skeletal and muscular system design, shown in Fig.~\ref{fig:skinning_oriented_robot_development}, carefully constrains the control points' B6DOF motions to match the target LBS-based motion space. This tendon-driven approach allows for complex, coordinated movements that closely mimic the desired facial expressions. A silicone mask, iteratively designed with considerations for shape, hardness, and thickness, serves as the facial skin. The mask's soft silicone provides exceptional flexibility, enabling the robot to achieve desired motions with high fidelity, effectively translating the LBS-based design into physical form.

\begin{figure*}[!t]
    \centering
    \includegraphics[width=\linewidth]{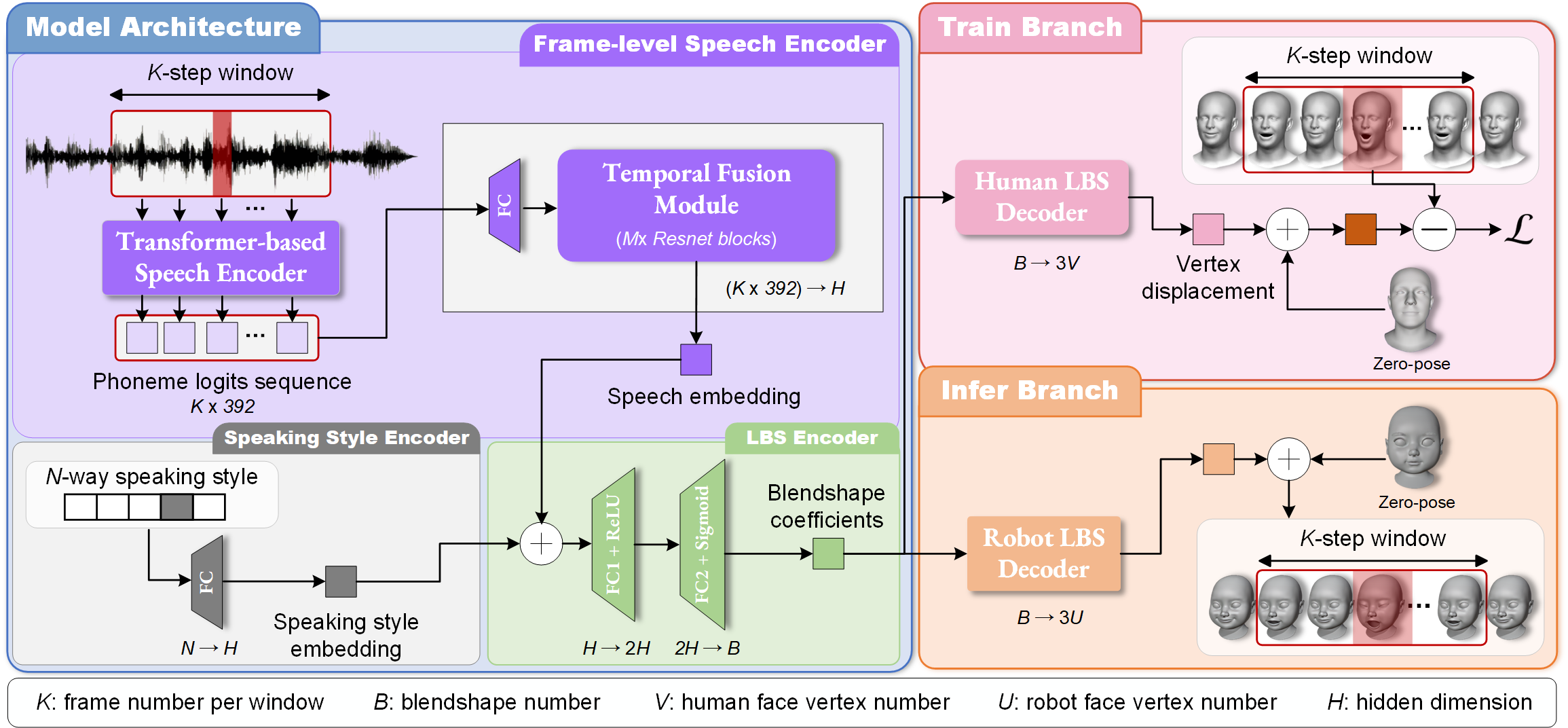}
    \caption{\textbf{The proposed speech-driven facial skinning motion imitation learning method.} The model architecture (blue section) comprises three key components: (1) a \textit{frame-level speech encoder} that processes audio input and generates phoneme logits, (2) a \textit{speaking style encoder} that captures individual speaking styles, and (3) an \textit{LBS encoder} that generates blendshape coefficients. During training (red section), the model learns to imitate human facial skinning motions by minimizing the difference between generated and target expressions. In the inference branch (orange section), the trained model generates blendshape coefficients for the robot LBS decoder, producing robot-specific facial skinning motions as reference signals for the downstream kinematics simulator.}
    \label{fig:imitation_learning}
    \vspace{-10pt}
\end{figure*}

\textbf{Electrical Design:} The electrical design aims to enable real-time skinning motion tracking by the animatronic face. To achieve responsive $25\mathrm{Hz}$ control for robot orchestration, the system must actuate each control point across its full motion range within $40\mathrm{ms}$.

To meet these stringent real-time requirements, specific components are carefully selected based on their capabilities and performance characteristics. The ESP32 microcontroller is chosen for its efficient on-chip signal processing capabilities, enabling real-time control. Two PCA9685 boards provide interfaces for driving the total $31$ servo motors required for articulation. Three high-torque DS3115 servos are selected for the neck module due to their ability to enable rapid rotations under the payload of the robot head, while $28$ lower-torque LFD-01M servos optimize cost while still providing adequate speed for the finer facial motions. With a $0.12\mathrm{sec}/60^{\circ}$ at $4.8\mathrm{V}$, the servos exceed the $40\mathrm{ms}$ actuation constraints when powered by a $5\mathrm{V}$ lithium battery.

The high-level system uses a GPU-accelerated PC to generate motion references from the learned model and run the simulator to solve inverse kinematics, generating servo commands for the ESP32 microcontroller. This setup ensures that the complex LBS-based motions are accurately translated into physical movements in real-time.

\section{Skinning Motion Imitation Learning}\label{sec:imitation_learning}

The goal is to learn a function $f(\cdot)$ that maps input speech $\mathbf{s}$ to blendshape coefficients $\boldsymbol{\theta}$, i.e. $f(\mathbf{s}) \rightarrow \boldsymbol{\theta}$, where $\boldsymbol{\theta} \in \mathbb{R}^B$ and $B$ is the number of blendshapes. The function $f(\cdot)$ is learned from a dataset $\mathcal{D}$ containing paired examples of speech and corresponding 3D human skinning motions. The learned function should generalize to unseen speech inputs and generate realistic, expressive facial motions.

Fig.~\ref{fig:imitation_learning} presents the proposed facial skinning motion imitation learning method, consisting of two key branches: (1) The training branch develops a model to generate LBS-based facial skinning motions, represented by blendshape coefficients, from input speech. This model is learned from 3D human demonstrations showing linkages between speech and facial skinning motions. (2) The inference branch leverages the robot LBS decoder to transform predicted blendshape coefficients into corresponding skinning motion reference signals. These signals further drive through the robot kinematics simulator to produce lifelike facial articulation.

\subsection{Model Architecture}\label{subsec:model_arch}

The model architecture $f(\cdot)$ takes raw speech waveforms $\mathbf{s}$ and generates blendshape coefficients $\boldsymbol{\theta}$ representing skinning motions. It contains: (i) a frame-level speech encoder extracting embeddings via a transformer-based phoneme logit extractor and temporal fusion module, (ii) a speaking style encoder embedding conditioning vectors, and (iii) an LBS encoder projecting conditioned speech embeddings to blendshape coefficients in the range $[0, 1]$. In this context, speaking style refers to the characteristic facial motion patterns exhibited by different speakers during speech production. These patterns encompass individual variations in articulation, facial muscle engagement, and expressive tendencies, which contribute to the unique visual signature of each speaker's facial movements during speech.

Specifically, the phoneme extractor follows the state-of-the-art transformer-based self-supervised pre-trained speech model, Wav2vec2~\cite{baevski2020wav2vec}, finetuned on the CommonVoice dataset~\cite{ardila2019common} with $53$ languages and $392$ phoneme classes to enhance cross-language generalization. It outputs a phoneme logits sequence at $49\mathrm{Hz}$ followed by resampling to $25\mathrm{Hz}$ to match the robot orchestration rate. The temporal fusion module progressively fuses $K$-step neighborhood logits over a sliding window into frame-level speech embeddings, comprising $M=\log_2 K$ stacked Resnet blocks with a kernel size of $3$, stride of $2$, and $H$ filters. The $N$-way speaking style vector accounts for cross-subject variations and is embedded with a fully-connected (FC) layer having $H$ hidden units. This component is crucial for capturing and reproducing individual speaking styles, enhancing the model's ability to generate diverse and personalized facial expressions. The LBS encoder takes the late-fused conditioned speech embedding as input and outputs blendshape coefficients with two successive layers: an FC layer having $2H$ hidden units with a ReLU activation, and an FC layer with $H$ hidden units followed by a Sigmoid activation.

\subsection{Training}\label{subsec:training}
The human LBS decoder transforms blendshape coefficients into vertex displacements. It is an FC layer with zero biases and a linear activation. Its weights are set and frozen using designed blendshapes over vertex displacements.

The loss function compares predicted and ground-truth vertices, regressing positions over the entire face. An additional weighted mouth term encourages improved lip synchronization, as mouth shapes strongly correlate with speech. For each predicted frame-level vertex, $\mathbf{\hat{y}}=\{\hat{y}_i\}_{i\in[0, V]}$, the model is trained by minimizing the loss $\mathcal{L}$ compared to the ground truth vertex, $\mathbf{y}=\{y_i\}_{i\in[0, V]}$. The loss is
\begin{equation}
    \mathcal{L} = \sum_{i=1}^{V}\left(y_i - \hat{y}_i\right)^2 + w_m\sum_{j=1}^{V_m}\left(y_j - \hat{y}_j\right)^2,
    \label{eq:loss}
\end{equation}
where $V_m$ is the number of vertices in the masked mouth region and $w_m$ is the mouth weight.

\begin{figure*}[!t]
    \centering
    \includegraphics[width=\linewidth]{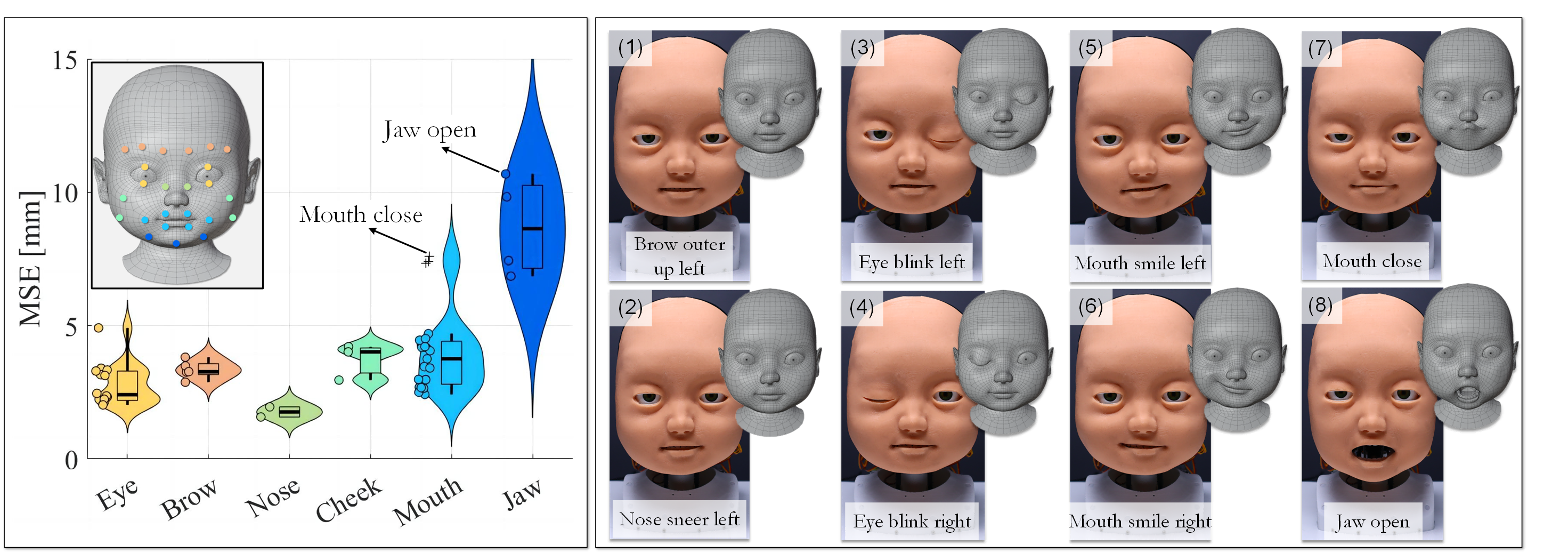}
    \caption{\textbf{Motion Space Validation.} \textbf{Actuated blendshape error for different facial regions (left figure):} Color-coded skinning landmarks represent different facial regions for evaluation. The 3D face model shows color-coded landmarks for different facial areas. Error distributions between simulated and physically actuated blendshapes are visualized using violin plots, box plots, and scattered points. Each point represents a single blendshape, evaluated using region-specific landmarks. Median errors (in mm) are provided for each facial region, ranging from $1.76\mathrm{mm}$ (nose) to $8.63\mathrm{mm}$ (jaw). \textbf{Qualitative comparison (right figure):} Visual comparison of eight simulated (gray mesh) versus actuated (realistic skin) blendshapes are shown. Blendshapes (1)-(6) demonstrate high accuracy, while (7) \textit{mouth close} and (8) \textit{jaw open} highlight limitations in the current design, exhibiting maximum errors for their respective regions.}
    \label{fig:exp1_motion_space}
    \vspace{-10pt}
\end{figure*}

\subsection{Inference}\label{subsec:inference}

The robot LBS decoder, a crucial component in the inference branch as illustrated in Fig.~\ref{fig:imitation_learning}, converts generated blendshape coefficients into skinning motions. Utilizing pre-designed robot blendshapes, this decoder operates with fixed weights established during the facial motion space design phase. While structurally analogous to the human LBS decoder, its weights are determined by applying designed robot's blendshapes to vertex displacements. The non-trainable nature of this decoder ensures consistent interpretation of blendshape coefficients throughout inference.

During the inference process, the trained model first generates a sequence of blendshape coefficients from input speech. This sequence undergoes temporal smoothing via a low-pass Butterworth filter to enhance stability. The robot LBS decoder then processes this filtered sequence, computing the skinning motion that serves as a reference signal for the downstream kinematics simulator. This approach maintains a consistent mapping between blendshape coefficients and resulting skinning motions, preserving the integrity of the designed facial expression space throughout inference.

\subsection{Implementation Details}\label{subsec:implementation_details}
The VOCASET dataset~\cite{cudeiro2019capture}, comprising audio-3D scan pairs of English utterances, is utilized for training and testing. This dataset was selected due to its high-quality 3D facial motion captures, diverse subjects, and phonetically balanced utterances, making it suitable for learning speech-driven facial expressions. VOCASET contains $255$ unique sentences shared among $12$ subjects, with $480$ facial motion sequences captured at $60\mathrm{Hz}$, each lasting $3$ to $4$ seconds. The 3D face mesh consists of $5023$ vertices.

For this study, VOCASET motions are resampled to $25\mathrm{Hz}$ to align with the robot orchestration rate. Subject-agnostic sequences are derived by subtracting zero-pose subject-specific face meshes from vertex sequences and adding them to the designed neutral (zero-pose) human face, $\mathbf{T}^H$. The data is split $80/10/10$ into train/validation/test sets, with two subjects reserved for testing generalizability. While VOCASET provides a foundation for initial model validation, the limited subject count may constrain generalization. This study serves as a proof-of-concept for the proposed approach, with exploration of larger, diverse datasets to enhance model generalizability being a potential area for future research.

The model is implemented in PyTorch with the following parameters: sliding window size $K=8$ ($320\mathrm{ms}$), hidden dimension $H=64$, and robot specifications of $U=4792$ vertices and $B=51$ blendshapes. Training utilizes the Adam optimizer with $\beta_1=0.9$, $\beta_2=0.99$, learning rate $1e^{-4}$, and weight decay $1e^{-4}$. The motion weight $w_m$ is set to $1$. Dropout layers with a rate of $0.1$ are inserted after each fully connected (FC) layer, except for the one with Sigmoid activation. The model is trained for $200$ epochs on an NVIDIA 4090 GPU, taking approximately $3\mathrm{sec}/\mathrm{epoch}$. The Wav2vec2-XLSR53 component remains fixed during training. For inference, a $7\mathrm{Hz}$ $5$th-order low-pass Butterworth filter is applied to smooth the output. The model demonstrates high-speed performance, generating blendshape coefficients from speech at a rate exceeding $4000\mathrm{Hz}$ on GPU.

\section{Experiments}\label{sec:exp}

\subsection{Robot Development Experiments}\label{subsec:robot_develop_exp}

\begin{figure*}[!t]
    \centering
    \includegraphics[width=\linewidth]{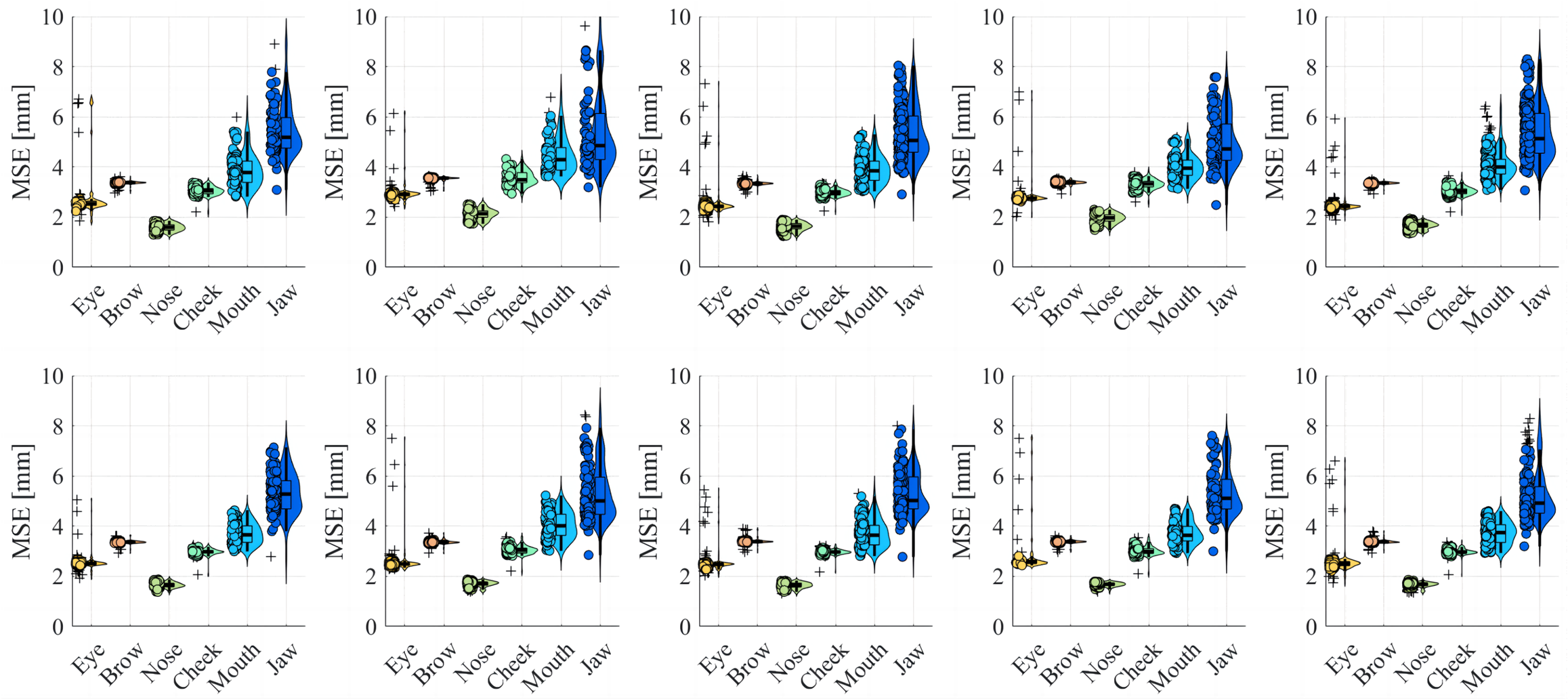}
    \caption{\textbf{Tracking Performance Validation.} MSE error distributions between simulated and physically actuated facial articulation sequences are presented using violin plots, box and whisker plots, and scattered points, with each point representing one frame. Evaluation landmarks are grouped by facial region. Ten realistic facial articulation sequences from different speakers with distinct speaking styles were evaluated. Mean median errors across the ten sequences for each facial region are $2.56\mathrm{mm}$ (eye), $3.39\mathrm{mm}$ (brow), $1.74\mathrm{mm}$ (nose), $3.08\mathrm{mm}$ (cheek), $3.86\mathrm{mm}$ (mouth), and $5.03\mathrm{mm}$ (jaw). This comprehensive visualization demonstrates the animatronic face's ability to maintain consistent tracking performance across diverse speaking styles while highlighting region-specific variations in accuracy.}
    \label{fig:exp2_tracking}
    \vspace{-10pt}
\end{figure*}

The experiments in this subsection aim to validate the proposed animatronic face in achieving the designed motion space and dynamic tracking performance. These experiments provide crucial evidence for the effectiveness of the skinning-oriented design approach and electro-mechanical design in enabling realistic and responsive facial expressions.

\textbf{Motion Space Validation:} This experiment quantitatively validates the physical realization of the designed LBS motion space defined by the robot blendshapes. Achieving the comprehensive set of blendshapes is essential for enabling expressive facial articulation. A VICON motion capture system is employed to track the physical skinning deformations. Reflective markers are attached to key control points and auxiliary skinning locations, corresponding to landmarks in the simulator. By individually actuating each blendshape and comparing the VICON marker positions with the simulated landmarks, the discrepancies can be measured using the mean squared error (MSE).

The results, presented in Fig. \ref{fig:exp1_motion_space}, demonstrate highly accurate realization of the designed motion space. Across various facial regions, the MSE errors consistently remain at the millimeter scale. For instance, the mouth blendshapes exhibit a median error of $3.76 \mathrm{mm}$, while the eye blendshapes show a median error of $2.41 \mathrm{mm}$. These low errors confirm the successful translation of the designed blendshapes into physical skinning deformations, validating the effectiveness of the skinning-oriented design approach.

\textbf{Tracking Performance Validation:} This experiment evaluates the animatronic platform's dynamic tracking performance in real-time by assessing its ability to follow reference facial motions. Achieving responsive and accurate tracking is crucial for enabling lifelike and synchronous facial expressions. The experimental setup employs the same VICON system, with markers attached to the robot's control points and auxiliary skinning locations. A diverse set of $10$ realistic facial articulation sequences from different speakers with distinct speaking styles from the training set, enhanced with random blinks, serve as reference signals. These sequences were carefully selected to encompass a wide range of full facial dynamics, representative of the system's intended operating conditions and potential real-world applications.

As demonstrated in Fig.~\ref{fig:exp2_tracking}, the animatronic platform exhibits remarkable tracking performance. The MSE errors between the reference signals and the tracked positions remain consistently low over time across all facial regions and speaking styles. Comparing these dynamically achieved errors with those obtained statically during the Motion Space Validation experiment presented in Fig. \ref{fig:exp1_motion_space}, the errors for different facial regions are very similar, on the millimeter scale. These results validate the effectiveness of the proposed electro-mechanical design in enabling precise and responsive facial motion control, even under dynamic conditions mimicking natural speech production.

\subsection{Imitation Learning Experiments}\label{subsec:imitation_learning_exp}

The following experiment assesses the validity of the proposed speech-driven motion synthesis method in generating realistic and expressive robot skinning motion references. Evaluating motion quality poses challenges due to the complex relationship between speech and facial expressions. As various plausible motions can match the same utterance, metrics such as prediction error are ineffective for assessing synthesis quality. Instead, a blind user study is employed to perform a perceptual evaluation and gauge the naturalness of the generated motions.

The user study involves a binary comparison between test sequences serving as ground truth and the developed model's output conditioned on all training subjects. The test sequences are distinct from the training and validation sets and involve subjects not used during training. These sequences, originally processed as dense human facial skinning, are projected onto the human blendshape basis by solving a constrained linear optimization problem. The resulting blendshape coefficients are then retargeted to the robot's facial skinning to obtain the ground-truth motions.

Both ground-truth and generated motions are rendered as textureless videos to focus the evaluation solely on motion quality. Participants are asked to choose the more natural motion matching the speech or indicate similar quality. The display order is randomized to prevent bias. Participants must pass a qualification test to ensure meaningful results.

In total, $12$ qualified participants evaluated $80$ video pairs three times each. Results show participants preferred generated motions in $29.2\%$ of tests, ground truth in $45.0\%$, and similar quality in $25.8\%$. These findings validate the motion synthesis model's effectiveness in generating natural, expressive robot skinning motions from speech, satisfying human perception.

The overall performance, including the motion synthesis on the developed animatronic robot face, is demonstrated in the supplementary video. The developed system is capable of automatically generating appropriate and dynamic facial expressions from speech in real-time, indicating the validity of the proposed skinning-centric approach that tightly integrates embodiment design and motion synthesis. However, it's crucial to acknowledge the study's limitations, such as the lack of in-person interactions with the physical robot and the limited participant pool. Future work should address these constraints to further validate the system's effectiveness in real-world scenarios.

\section{Conclusions and Future Works}\label{sec:conclusion}
This paper introduces a novel, skinning-centric approach to drive animatronic robot facial expressions from speech. By employing LBS as the core representation, tightly integrated innovations in embodiment design and motion synthesis are achieved. LBS serves as a guiding principle for the actuation topology, enables seamless human expression retargeting, and facilitates speech-driven facial motion generation. Key technical contributions include the skinning-oriented design approach, the electro-mechanical implementation achieving millimeter-scale accuracy, and the imitation learning method for motion synthesis. Experimental results conclusively demonstrate that the developed animatronic robot face successfully generates highly realistic facial expressions from speech automatically and in real-time, validating the effectiveness of the proposed approach.

Future research can build upon this work in two primary directions. First, a general robot facial muscular system could be explored that drives any human-like robot face simply by replacing the customized skull and skinning components, facilitating easy fabrication and control of customized animatronic faces. This could involve developing modular actuation units and adaptive control algorithms to accommodate various facial structures. Second, more advanced imitation learning methods could empower speech-synchronized robot facial expressions with controllable emotions to enable complex facial signals that support richer, more engaging human-robot interactions. This might entail incorporating multi-modal inputs beyond speech and developing more sophisticated emotion models. As animatronic robot technology continues to mature, realistic robotic faces will likely become commonplace across entertainment, education, healthcare, and other facets of life. This pioneering skinning-based approach establishes a strong foundation for that future.

\textbf{Acknowledgement:} This work was supported in part by the National Natural Science Foundation of China (No.62376031).

{\small
\bibliographystyle{ieeetr}
\bibliography{reference}
}

\end{document}